\documentclass[sigconf,nonacm]{acmart}

\usepackage{graphicx}

\AtBeginDocument{%
  }

\usepackage{water-curtain-cave}

\begin{document}

\newcommand{\rocs}{ROCS\xspace}
\newcommand{\rocsified}{ROCSified\xspace}
\newcommand{\meta}{a major Internet company\xspace}

\newcommand\fixme[1]{{\bf \color{red} #1 }}
\newcommand\yxedit[1]{{\bf \color{blue} #1 }}
\newcommand\lledit[1]{{\bf \color{pink} #1 }}

\newcommand{\hy}[1]{{\color{red} #1}}
\newcommand{\cq}[1]{{\color{cyan} \{CQ: #1}\}}

\title{\rocs: Request-Oriented Compute Sharing for Efficient Large-Scale Recommendation}

\author{%
Yuxin Chen\textsuperscript{*},
Liang Luo\textsuperscript{*},
Buyun Zhang\textsuperscript{*},
Jian Jiao\textsuperscript{*},
Boda Li\textsuperscript{*},
Haoyu Wang\textsuperscript{*},
Tongyi Tang, Ao Cai, Zijian Shen, Zhengkai Zhang, Wenyi Xie, Ryan Dick, Han Liu, Neng Shi,
Bin Yu, Jianbo Xiao, Shuyao Bi, Hongtao Yu, Yuanwei Fang, Zhuoran Zhao, Sijia Chen,
Yang Chen, Shuqi Yang, Qianru Li, Zikun Liu, Wei Ling, Sihan Zeng, Longhao Jin, Jiaxin Lu,
Yinbin Ma,
Jiawei Li, Yichen Ruan, Yong Ler Lee, Birmingham Guan, Zijian Li, Jianbo Sun,
Zhengyu Zhang, Zeliang Chen, Xiaohan Wei, Yuchen Hao, GP Musumeci,
Venkatesh Ranganathan, Yantao Yao, Chunqiang Tang, Wenlin Chen, Santanu Kolay, Ellie Wen
}
\email{\{yuxinc,liangluo,buyunz,jianj,liptds,dwen\}@meta.com}
\affiliation{%
  \institution{Meta AI}
  \city{Menlo Park}
  \state{California}
  \country{USA}
}

\renewcommand{\shortauthors}{ROCS Team}

\begin{abstract}
Modern recommendation models gain prediction quality by scaling feature-interaction and sequence modules, but production cost constraints cap how far systems can scale.

In this work, we propose \textit{Request-Oriented Compute Sharing (ROCS)}, a modeling and inference paradigm that exploits a unique property of recommendation inference: each user request is evaluated against many candidates, while request-side features are shared across candidates.
ROCS defers request--candidate interactions as late as possible, isolates candidate-dependent representations, and evaluates substantial portions of the model once per request rather than once per candidate, significantly improving inference efficiency while maintaining or improving prediction quality.
To realize this paradigm, we develop \textit{Generalized Layer Masking (GLM)} to enforce candidate isolation in feature-interaction architectures, and \textit{Deep Cross Attention (DCA)} to extend request-oriented sharing to sequence architectures. To support efficient GPU deployment, we co-design \textit{In-Kernel Broadcast Optimization (IKBO)} that significantly accelerates ROCS model execution~\footnote{Code has been open-sourced at \href{https://github.com/pytorch/FBGEMM/tree/main/fbgemm_gpu/experimental/ikbo}{this repository}.}.

Experiments on public benchmarks show that ROCS consistently improves the quality--efficiency tradeoff across recommendation backbones. On production-scale workloads, ROCS achieves up to a 3$\times$ QPS improvement on retrieval models without quality degradation and a 0.5\% relative LogLoss improvement with a 50\% QPS gain on a short-form video ranking model.
ROCS has been deployed across large-scale recommendation systems at Meta, spanning ads and organic surfaces, retrieval and ranking stages, and more than two orders of magnitude in inference complexity, delivering significant online gains at reduced infrastructure cost.
\end{abstract}

\maketitle

\renewcommand{\thefootnote}{\fnsymbol{footnote}}
\footnotetext[1]{Equal contribution.}
\renewcommand{\thefootnote}{\arabic{footnote}}

\section{Introduction}

\begin{figure*}[t]
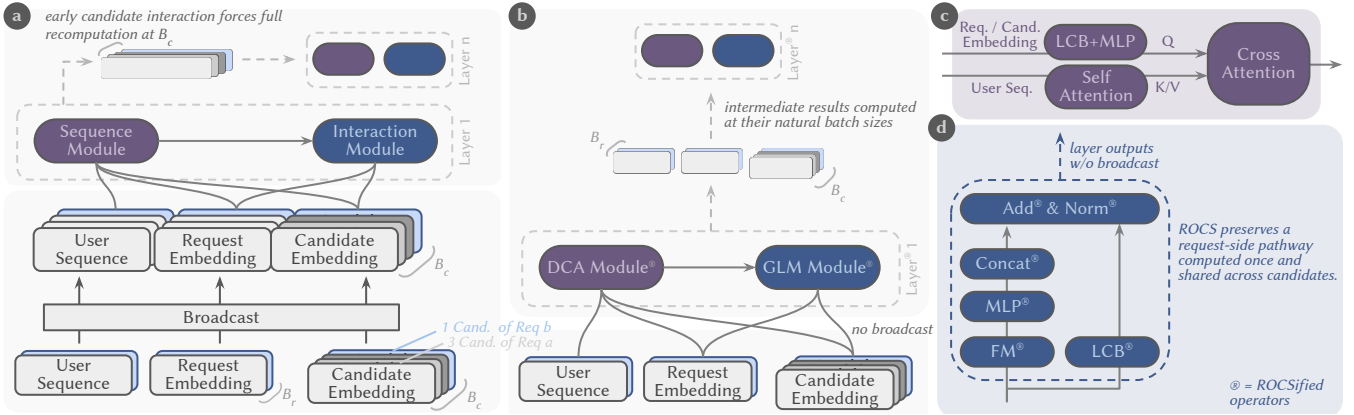

    \rocsfigure{5}{2.15in}
    \vspace{-0.3in}
    \caption{Overview of ROCS.
(a) Conventional models broadcast request-side inputs and introduce request–candidate interactions early, causing substantial redundant computation across candidates.
(b) ROCS maintains a reusable request-side pathway by ROCSifying operators to satisfy a dependency contract and eliminate redundant computation.
(c) Deep Cross Attention (DCA) extends ROCS to sequence modeling through deep candidate-conditioned retrieval over shared sequence representations.
(d) Detailed view of Generalized Layer Masking (GLM) applied to the Wukong backbone.
}
    \label{fig:rocs_overview}        
\end{figure*}

Recommendation scaling laws~\cite{zhang2024wukong, zhu2025rankmixer} show that prediction quality improves predictably with model compute, but production capacity and latency constraints cap the compute budget per request. Real-world systems must therefore carefully navigate the tension between predictive accuracy and serving cost.

Prior work has largely explored system-level approaches to alleviate these constraints. Methods like model unification~\cite{lattice} reduce overall computation by consolidating specialized models, while knowledge transfer~\cite{solaris, exlargefm} offloads heavy computation to asynchronous pipelines and distills knowledge into lightweight serving models. While effective, these methods are limited by domain interference, distillation inefficiency, and freshness requirements~\cite{deck}.  

In this work, we attack this problem directly via a modeling-driven approach. 
Our core insight leverages an inherent property of recommendation inference: a single user request is evaluated against many candidate items.
For a given request, request-side features remain identical across candidates, whereas candidate-side features vary.
Therefore, if request--candidate interaction is delayed, request-side computation can be evaluated once and reused across all associated candidates.
Because computation before the interaction point is amortized across candidates, whereas computation after it is paid per candidate, maximizing delayed interaction directly increases the amount of reusable computation.

Unfortunately, conventional ranking architectures mandate early fusion of request and candidate to extract cross-features~\cite{zhang2024wukong, dlrm, zhang2022dhen, wang2021dcn, dmt}. 
This is compounded in modern sequence modeling: when long user sequences are early-fused with candidate signals~\cite{zeng2025interformer, hou2026kunlun}, quadratic-cost sequence operators and their downstream computation must be executed separately for every candidate, making real-time serving increasingly impractical.
At the opposite extreme, two-tower architectures~\cite{cdeep} maximize reuse by deferring request--candidate interaction to a final scoring function, but sacrifice intermediate interaction and model expressiveness.

Recent transformer-based and architecture-specific approaches~\cite{lu2026computeonceugseparationefficient, zhang2026onetransunifiedfeatureinteraction, guo2025requestonlyoptimizationrecommendationsystems} explore a middle ground by preserving reusable request-side representations through masked interactions.
However, these designs are tied to particular model structures and do not govern the dependencies introduced by other substantial components in practical recommendation models.

We identify request-level compute redundancy as a structural mismatch between recommendation model architectures and request-to-candidate execution.
To address this mismatch, we propose \textit{Request-Oriented Compute Sharing (ROCS)}, a modeling and inference paradigm that systematically exposes, preserves, and efficiently executes request-shared computation.
ROCS maintains a candidate-independent pathway while allowing candidate-side representations to consume request-side signals throughout the model.
This asymmetric dependency structure delays candidate information from entering shared representations without eliminating intermediate request--candidate interaction.
As a result, substantial portions of the model can be evaluated once per request and reused across all associated candidates.
ROCS consists of four components:

\begin{packed_itemize}
  \item \textbf{Generalized Layer Masking (GLM)} defines a compositional operator-level dependency contract that preserves reusable request-side computation across feature-interaction modules.
  \item \textbf{Deep Cross Attention (DCA)} extends ROCS to sequence models with deep candidate-conditioned retrieval. %
  \item \textbf{Request-Oriented Resource Reallocation (RRR)} reinvests computation savings to improve the quality--efficiency frontier.
  \item \textbf{In-Kernel Broadcast Optimization (IKBO)} efficiently executes ROCS models without materializing broadcast.
\end{packed_itemize}

Across public benchmarks and production-scale workloads, ROCS consistently improves the quality--efficiency frontier across recommendation backbones, achieving up to 3$\times$ the baseline QPS on retrieval without quality degradation and a 0.5\% relative LogLoss improvement with a 50\% QPS gain on short-form video ranking.
ROCS has been deployed across ads and organic systems spanning retrieval and ranking over two orders of magnitude in inference complexity, delivering online gains at reduced infrastructure cost.

\section{Related Work}

\para{Large Recommendation Models.}
Modern recommendation models predict user actions from heterogeneous inputs, including request-level context such as time and location, user behavior sequences, and candidate-level features such as item category, as illustrated in Figure~\ref{fig:rocs_overview}.
Sparse and dense features are typically embedded or transformed before being processed by feature-interaction and sequence modules.
Recent work has shown that prediction quality improves with increased compute in both feature-interaction architectures~\cite{zhang2024wukong,zhu2025rankmixer,jiang2026tokenmixerlargescalinglargeranking,guo2023embedding,ardalani2022understanding} and sequence architectures~\cite{zhai2024actionsspeaklouderwords,chai2025longer,ding2026ultrahstu}, exhibiting scaling behavior analogous to that observed in language models~\cite{kaplan2020scaling}.
More recent architectures jointly model sequential and non-sequential features throughout the network~\cite{zeng2025interformer,huang2026hyformerrevisitingrolessequence,zhang2026onetransunifiedfeatureinteraction,huang2026mixformercoscalingdensesequence,hou2026kunlun}.
Although this design strengthens candidate-conditioned interaction, it also increases the computation required to score each candidate.

\para{Efficient LRM Serving.}
The growing computational cost of large recommendation models has motivated specialized techniques such as model unification~\cite{lattice}, knowledge transfer and distillation~\cite{solaris,exlargefm}, and low-precision execution~\cite{luo2026lokalowprecisionkernelapplications}.
These approaches reduce serving cost by consolidating models, transferring computation to offline or asynchronous pipelines, or improving hardware efficiency.
Model unification and knowledge transfer may nevertheless be constrained by interference across tasks, imperfect transfer, and freshness requirements~\cite{deck}.
ROCS instead restructures the model to eliminate redundant per-candidate computation.
It is complementary to these approaches and can be combined with them.

\para{Request-Oriented Recommendation Models.}
Request-oriented designs separate request- and candidate-side computation and reuse the request-side subgraph across candidates.
Two-tower models encode users and items independently and defer their interaction to a final similarity function~\cite{cdeep}.
This structure enables efficient retrieval but restricts intermediate request--candidate interaction and is therefore less commonly used as the sole architecture for compute-intensive ranking stages.

Transformer-based recommendation models can also preserve separate request- and candidate-side token representations through attention masking~\cite{huang2026hyformerrevisitingrolessequence,guo2025requestonlyoptimizationrecommendationsystems,ding2026ultrahstu}.
Attention masking, however, does not govern the dependencies introduced by the substantial non-attention components in practical recommendation models, including MLPs, feature-compression layers, and explicit feature-interaction operators.
Once these components mix request- and candidate-side inputs, their results become candidate-dependent and cannot be reused exactly across candidates.
Moreover, architectures that inject candidate information into the shared sequence~\cite{zeng2025interformer,hou2026kunlun} make subsequent sequence representations candidate-dependent and prevent their exact reuse across candidates.

UGSEP~\cite{lu2026computeonceugseparationefficient} introduces a RankMixer-specific masked token-mixing design that preserves reusable user-side tokens while compensating for the resulting loss of interaction capacity.
In contrast, ROCS formulates request sharing as an operator-level dependency invariant that is closed under composition and applies across heterogeneous feature-interaction and sequence operators.
ROCS further co-designs the model structure with IKBO. IKBO keeps request- and candidate-side tensors at their natural batch sizes and resolves request--candidate association inside the consuming GPU kernels without materializing request-to-candidate broadcast.

\section{Design of ROCS}
\label{sec:rocs}

\rocs is guided by three design goals:
(1) it should apply broadly to existing recommendation architectures.
(2) it should defer request--candidate interaction to maximize reusable request-side computation.
(3) it should preserve or improve prediction quality despite delayed interaction.
To achieve these goals, \rocs transforms existing models into architectures with an explicit reusable request-side subgraph. Its design consists of three techniques:
\begin{packed_itemize}
  \item \textbf{Generalized Layer Masking (GLM)} defines an operator-level dependency contract and a compositional transformation that exposes reusable computation across heterogeneous feature-interaction modules.
  \item \textbf{Deep Cross Attention (DCA)} extends request-oriented sharing to sequence models through shared sequence encoding and deep candidate-conditioned retrieval.
  \item \textbf{Request-Oriented Resource Reallocation (RRR)} reinvests the savings from request-side amortization to improve the quality--efficiency frontier.
\end{packed_itemize}

\begin{figure*}[t]
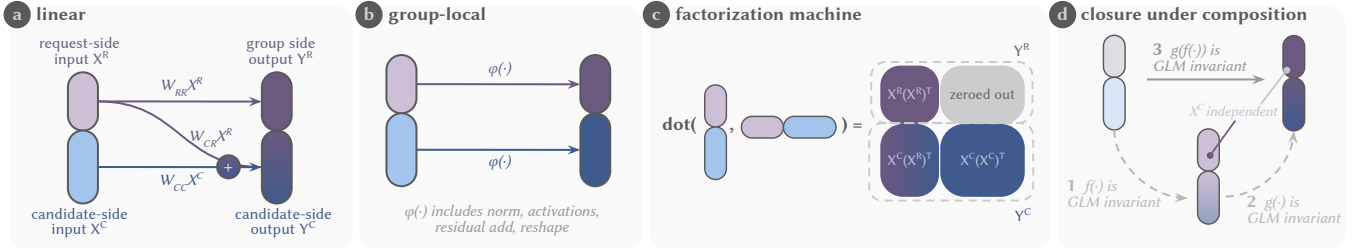

    \rocsfigure{6}{1.3in}
    \vspace{-2em}
    \caption{Construction of GLM modules, illustrated with two-group scenario where group 0 is request-side and group 1 is candidate-side. (a) linear and linear compression. (b) group-local operators. (c) factorization machine. (d) GLM invariant is preserved under composition.}
    \label{fig:glm}        
\end{figure*}

\subsection{Generalized Layer Masking}
\label{sec:rocs:glm}

For a request evaluated against several candidates, an intermediate request-side result is reusable only if changing the candidate cannot change that result. Generalized Layer Masking (GLM) makes this requirement an explicit contract for every transformed operator in the feature-interaction stack. The contract is inspired by causal attention masks~\cite{vaswani2017attention}, but its implementation is specific to each operator. GLM cannot generally be enforced by masking an operator’s output: once the operator has mixed request- and candidate-side inputs, the retained request-side values may already depend on the candidate. The dependency restriction must therefore be enforced inside each operator.

\noindent\textbf{GLM invariant.}
Let input $x$ be partitioned into $K$ ordered groups $x^{(0)},\ldots,x^{(K-1)}$. Let $f^{(i)}(x)$ denote output group $i$, using the same group order. For fixed model parameters and a deterministic forward pass, we call $f$ \emph{GLM-compliant} (or \emph{\rocsified}) if, for every group $i$ and every pair of inputs $x,x'$,
\begin{equation}
x^{(0:i)}=x'^{(0:i)}
\quad\Longrightarrow\quad
f^{(i)}(x)=f^{(i)}(x'),
\label{eq:glm-invariant}
\end{equation}
where $x^{(0:i)}=(x^{(0)},\ldots,x^{(i)})$. Equivalently, there is a function $\bar f_i$ such that
$f^{(i)}(x)=\bar f_i(x^{(0)},\ldots,x^{(i)})$. Thus information may flow from an earlier group to a later group, but not in the reverse direction. This is a functional dependency constraint, not an assumption that the feature groups are statistically independent.

\rocs normally uses two groups. We denote request-side group 0 by $x^R$ and candidate-side group 1 by $x^C$. Equation~\eqref{eq:glm-invariant} then requires
\begin{equation}
f(x^R,x^C)
=\bigl(f^R(x^R),\,f^C(x^R,x^C)\bigr).
\label{eq:glm-two-groups}
\end{equation}
The request-side output is invariant to the candidate. The candidate-side output can still use both inputs and model their interaction.

The group-level dependency pattern is the block lower-triangular mask:
\begin{equation}
M_{ij}=\mathbf{1}\{j\leq i\}.
\label{eq:glm-mask}
\end{equation}
Here $M$ records which input-output group pairs are allowed; each operator enforces this pattern according to its structure. For a matrix $X\in\mathbb{R}^{n\times d}$, row block $i$ has shape $X^{(i)}\in\mathbb{R}^{n_i\times d}$, with $n=\sum_i n_i$. For a vector $x\in\mathbb{R}^m$, coordinate block $i$ has shape $x^{(i)}\in\mathbb{R}^{m_i}$, with $m=\sum_i m_i$. Input and output groups follow the same group order, although their corresponding group sizes may differ

\subsubsection{Linear and Linear Compression Layers}
Consider an affine layer with grouped input
$x=[x^{(0)};\ldots;x^{(K-1)}]$ and grouped output
$y=[y^{(0)};\ldots;y^{(K-1)}]$. Here
$x^{(j)}\in\mathbb{R}^{m_j^{\mathrm{in}}}$ is the vector in input group $j$, and
$y^{(i)}\in\mathbb{R}^{m_i^{\mathrm{out}}}$ is the vector in output group $i$.
Weight block $W_{ij}\in\mathbb{R}^{m_i^{\mathrm{out}}\times m_j^{\mathrm{in}}}$ maps input group $j$ to output group $i$, and $b^{(i)}\in\mathbb{R}^{m_i^{\mathrm{out}}}$ is its output bias. GLM masks the weight \emph{blockwise}:
\begin{equation}
\begin{aligned}
\widetilde W_{ij}&=M_{ij}W_{ij},\\
y^{(i)}&=\sum_{j=0}^{K-1}\widetilde W_{ij}x^{(j)}+b^{(i)} = \sum_{j=0}^{i}W_{ij}x^{(j)}+b^{(i)}.
\end{aligned}
\label{eq:glm-linear}
\end{equation}

An LCB applies the same construction to the row axis of an embedding matrix. Let $X^{(j)}\in\mathbb{R}^{n_j^{\mathrm{in}}\times d}$ and $Y^{(i)}\in\mathbb{R}^{n_i^{\mathrm{out}}\times d}$. With
$W_{ij}\in\mathbb{R}^{n_i^{\mathrm{out}}\times n_j^{\mathrm{in}}}$, the \rocsified LCB is
\begin{equation}
Y^{(i)}=\sum_{j=0}^{i}W_{ij}X^{(j)}.
\label{eq:glm-lcb}
\end{equation}
For the two-group case, its weight has the explicit form
\begin{equation}
\begin{bmatrix}Y^R\\Y^C\end{bmatrix}
=
\begin{bmatrix}W_{RR}&0\\W_{CR}&W_{CC}\end{bmatrix}
\begin{bmatrix}X^R\\X^C\end{bmatrix}.
\label{eq:glm-lcb-two-groups}
\end{equation}
The term $W_{CR}X^R$ is the request-derived contribution, which IKBO supplies without explicitly materializing the broadcast (Section~\ref{sec:ikbo}).

\subsubsection{Group-Local Operations}
An operator that aggregates coordinates, such as normalization, must compute the output of group $i$ without using values from any group after $i$. We use the stricter and simpler group-local form
\begin{equation}
y^{(i)}=\operatorname{Norm}_i(x^{(i)}),
\label{eq:glm-norm}
\end{equation}
which covers per-group LayerNorm or RMSNorm and inference normalization with fixed statistics. Statistics computed from concatenated groups can depend on a later group and violate Eq.~\eqref{eq:glm-invariant}.

Unary pointwise activations are GLM-compliant. For multiple inputs, we apply an elementwise operator $\phi$ separately to aligned, broadcast-compatible blocks:
\begin{equation}
y^{(i)}=\phi(x_1^{(i)},\ldots,x_q^{(i)}).
\label{eq:glm-elementwise}
\end{equation}
This covers activations, residual addition, and other pointwise merges. Concatenation and reshaping are also compliant when they preserve group labels.

\subsubsection{Factorization Machine}

A factorization machine explicitly computes pairwise interactions between input embeddings. To satisfy the GLM invariant, we apply the lower-triangular block mask, so that output group $i$ retains only interactions with input groups up to $i$:
\begin{equation}
\begin{aligned}
Y^{(i)}
=
\left[
M_{i,0}X^{(i)}(X^{(0)})^\top,
\ldots,
M_{i,K-1}X^{(i)}(X^{(K-1)})^\top
\right]\\
=
\left[
X^{(i)}(X^{(0)})^\top,
\ldots,
X^{(i)}(X^{(i)})^\top,
\mathbf{0},
\ldots,
\mathbf{0}
\right].
\label{eq:glm-fm}
\end{aligned}
\end{equation}
Thus, $Y^{(i)}$ depends only on $X^{(0)},\ldots,X^{(i)}$ and is GLM-compliant.

For the two-group case, the output takes the following form:
\begin{equation}
\begin{bmatrix}
Y^R\\
Y^C
\end{bmatrix}
=
\begin{bmatrix}
X^R(X^R)^\top & 0\\
X^C(X^R)^\top & X^C(X^C)^\top
\end{bmatrix}.
\label{eq:glm-fm-two-groups}
\end{equation}
The request-side output contains only request--request interactions, while the candidate-side output includes both request--candidate and candidate--candidate interactions.

\subsubsection{Closure under composition}
Let $f$ and $g$ be GLM-compliant with aligned intermediate groups. If two inputs agree through group $i$, compliance of $f$ makes its outputs agree through group $i$. Compliance of $g$ then makes output group $i$ agree. Hence $g\circ f$ is GLM-compliant. By induction, MLPs and interaction stacks composed of the constructions above preserve the invariant.

As a result, consider candidates $C_1,\ldots,C_N$ associated with the same request $R$. Every \rocsified module satisfies
\begin{equation}
\bigl[F(X^R,X^{C_1})\bigr]^R=\cdots=\bigl[F(X^R,X^{C_N})\bigr]^R=F^R(X^R).
\label{eq:glm-sharing}
\end{equation}
Its request-side path may therefore be evaluated once and reused across all $N$ candidates while preserving the deterministic output of the same \rocsified module under a broadcasted execution. This establishes the semantic validity of compute sharing.

\subsection{Deep Cross Attention}
\label{sec:rocs:dca}

Candidate-aware sequence retrieval is widely used in recommendation: the current candidate provides context for selecting relevant events from the user behavior sequence~\cite{zhou2018din,pi2020sim,chang2023twin,si2024twinv2,chai2025longer,zeng2025interformer,hou2026kunlun}.
Introducing candidate context earlier in sequence modeling can improve prediction quality~\cite{yan2022apg,lattice,zeng2025interformer,hou2026kunlun}, but makes the sequence candidate-dependent and requires its expensive operators and downstream computation to be repeated for every candidate.

\emph{Deep Cross Attention} (DCA) separates candidate-conditioned sequence modeling into shared sequence encoding and candidate-conditioned retrieval.
A request-only sequence encoder first produces a candidate-independent representation $S_i$ at each layer.
A GLM-compliant query generator then constructs request- and candidate-side queries, which retrieve information from $S_i$ through separate cross-attention operations.
Figure~\ref{fig:rocs_overview}(c) shows this design.

\para{Shared Sequence Encoding.}
Let $S_0$ be the embedded user sequence, and let
$S_i=\operatorname{Enc}_i(S_{i-1})$ denote its representation at layer $i$.
Because the sequence stream excludes candidate inputs, every $S_i$ and its K/V projections are computed once per request and shared across associated candidates.

\para{Benefiting from Depth.} 
DCA places cross-attention at every interaction layer, allowing later queries to retrieve complementary information from the sequence representation at that depth using richer request--candidate context.

\para{ROCSified Query Generation.}
Let $Y_{i-1}^R$ and $Y_{i-1}^C$ denote the request- and candidate-side outputs of GLM layer $i-1$.
DCA constructs the two query groups using ROCSified $\operatorname{LCB}_i$ and $\operatorname{MLP}_i$:
\begin{equation}
\begin{bmatrix}
Q_i^R\\
Q_i^C
\end{bmatrix}
=
\operatorname{MLP}_i
\left(
\operatorname{LCB}_i
\left(
\begin{bmatrix}
Y_{i-1}^R\\
Y_{i-1}^C
\end{bmatrix}
\right)
\right).
\label{eq:dca-query}
\end{equation}

\para{Group-wise Cross Attention.}
The two query groups attend to the shared sequence representation:
\begin{equation}
\begin{aligned}
A_i^R
&=
\operatorname{Attn}
\left(
Q_i^R,
S_iW_{K,i}^R,
S_iW_{V,i}^R
\right),\\
A_i^C
&=
\operatorname{Attn}
\left(
Q_i^C,
S_iW_{K,i}^C,
S_iW_{V,i}^C
\right).
\end{aligned}
\label{eq:dca-attention}
\end{equation}
The attention outputs are concatenated with the corresponding GLM representations before the next interaction layer.

\para{Preservation of the GLM Invariant.}
The shared sequence and request-side query depend only on request-side inputs, whereas the candidate-side query may depend on both request and candidate inputs.
Thus, request-side attention is computed once per request, while candidate-side attention remains candidate-specific.
DCA therefore shares sequence encoding and K/V projections while retaining candidate-conditioned retrieval.
It can be composed with the GLM modules in Section~\ref{sec:rocs:glm}.

\subsection{Request-Oriented Resource Reallocation}

GLM and DCA maximize computation sharing by constraining when candidate-specific information enters the model.
However, ROCSification is not necessarily expressiveness-preserving under a fixed model capacity.
For example, a conventional layer may devote its full output capacity to candidate-dependent representations, whereas its ROCSified counterpart reserves part of that capacity for reusable request-side representations.
Directly ROCSifying a model may therefore reduce prediction quality.

\emph{Request-oriented Resource Reallocation} (RRR) addresses this trade-off by reinvesting the saved computation into scaling the request-side pathway, following the scaling behavior of modern recommendation backbones.
Because each request is evaluated against multiple candidates, request-side computation is amortized across them, making request-side scaling substantially cheaper than equivalent candidate-side scaling.
This asymmetric cost structure provides a favorable lever for recovering or improving model quality while retaining the efficiency benefits of compute sharing.

We control this allocation using a request-side scaling ratio $r$, which determines the relative capacity assigned to request-side representations in each ROCS module.
A larger $r$ increases the capacity of the reusable request-side pathway.
Unless otherwise specified, we use the same $r$ across all ROCS modules.

\section{Efficient Execution of ROCS with In-kernel Broadcast Optimization}
\label{sec:ikbo}

ROCS restructures a recommendation model to expose a reusable request-side subgraph. A direct dense implementation may nevertheless expand request-side tensors to the candidate batch size by replicating each tensor for all associated candidates, a process we refer to as \emph{request-to-candidate broadcast}. Such replication increases memory consumption, introduces unnecessary computation, and incurs additional memory traffic.

We therefore implement ROCS using \emph{In-Kernel Broadcast Optimization} (IKBO), a GPU execution strategy specialized for the structured computation exposed by ROCS. IKBO stores request- and candidate-side tensors at their natural batch sizes, $B_r$ and $B_c$, respectively, and uses an index map $\mathbf{m}$ to associate each candidate with its corresponding request. Rather than materializing request-to-candidate broadcast, IKBO loads the corresponding request-side result on demand inside the candidate-side GPU kernel.

We apply this strategy to the two dominant operators in ROCS: linear operators in GLM and cross-attention operators in DCA. We first describe how ROCS computations are decomposed into compact request- and candidate-batched components, and then present the corresponding IKBO kernels for GLM and DCA.

\subsection{Decomposing \rocsified Operators}
\label{sec:decompose_rocs}

For efficient execution, a \rocsified operator can be separated into request-batched and candidate-batched components. The request-batched component is evaluated once per request and produces both the request-side output and any request-derived contribution required by the candidate side. A decomposed dense implementation would then broadcast this contribution to the candidate batch and combine it with the candidate-local computation. This decomposition removes redundant request-side computation, but it does not yet eliminate the materialized broadcast.

We use LCB as a concrete example. Consider the two-group LCB in
Eq.~\ref{eq:glm-lcb-two-groups}, with request- and candidate-side inputs
stored at their natural batch sizes:
$
\mathbf{X}^{R}
\in \mathbb{R}^{B_r \times n_r \times d},
\mathbf{X}^{C}
\in \mathbb{R}^{B_c \times n_c \times d}.
$
An index map associates candidate $i$ with request $m_i$:
$
\mathbf{m}
\in
\{0,\ldots,B_r-1\}^{B_c}.
$

Rather than first replicating $\mathbf{X}^{R}$ to the candidate batch, we compute the two request-dependent terms together in a single GEMM at batch size $B_r$:
\[
\begin{bmatrix}
\mathbf{Y}^R \\
\mathbf{Z}^{R\rightarrow C}
\end{bmatrix}
=
\begin{bmatrix}
W_{RR} \\
W_{CR}
\end{bmatrix}
\mathbf{X}^R,
\]
where
$\mathbf{Z}^{R\rightarrow C}=W_{CR}\mathbf{X}^R$
corresponds to the request-derived contribution in
Eq.~\ref{eq:glm-lcb-two-groups}. 
The candidate-side output is then
\begin{equation}
\mathbf{Y}^C_i
=
\mathbf{Z}^{R\rightarrow C}_{m_i}
+
W_{CC}\mathbf{X}^C_i.
\label{eq:broadcast_add}
\end{equation}

DCA admits an analogous decomposition. Let $\mathbf{S}^{R}$ denote the
request-side sequence representation stored at batch size $B_r$.
The linear K/V projections can be applied before logical request-to-candidate broadcast. 
For either $W \in \{W_K, W_V\}$,
\[
\left(\mathbf{S}^{R}_{m_i}\right)W
=
\left(\mathbf{S}^{R}W\right)_{m_i}.
\]
The projected K/V tensors are therefore computed once per request.
However, candidate $i$ has its own attention query and must attend to
the K/V tensors of its associated request $m_i$. A dense attention operator would therefore require candidate-batched K/V tensors.

Both decompositions compute request-side contributions once at batch size
$B_r$ and associate them with candidate-side computation through
$\mathbf{m}$. A conventional dense implementation would still materialize
these results at batch size $B_c$. IKBO instead resolves $\mathbf{m}$
inside the consuming GEMM and attention kernels, eliminating the
remaining broadcast and its associated memory traffic.

\subsection{In-Kernel Broadcast Optimization for GLM}

\begin{figure}[t]
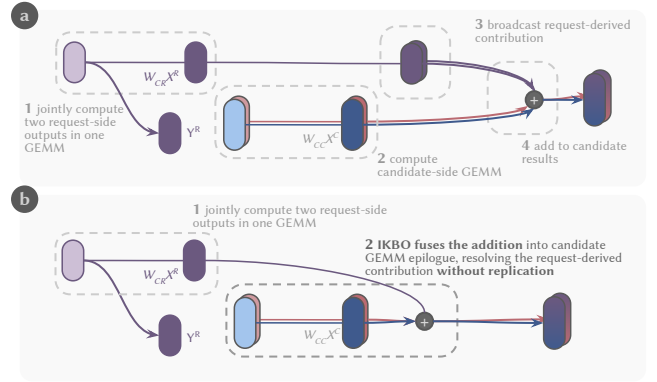

\rocsfigureonecol{10}{2in}
\vspace{-0.3in}
\caption{(a) Dense LCB implementation. (b) IKBO LCB implementation.}
\label{fig:ikbo_lcb}
\end{figure}

Decomposition removes redundant arithmetic, but an efficient implementation must also eliminate the data movement and scheduling overheads introduced by the split execution. IKBO addresses these overheads progressively. It first fuses broadcast-and-add into the candidate-side GEMM epilogue, then overlaps the heavier epilogue with computation through warp specialization. It further co-schedules request- and candidate-side tiles in a single kernel and repairs alignment inefficiencies introduced by decomposition.

\subsubsection{Fusing Broadcast and Add into GEMM Epilogue}
After decomposing a \rocsified operator, the bottleneck shifts to the memory-bound broadcast-and-add operation in Eq.~\eqref{eq:broadcast_add}. In a naive implementation, the candidate-side GEMM output is first written to HBM. It is then read back and combined with the corresponding precomputed request-side contribution, after which the final candidate-side output is written to HBM. This execution materializes an unnecessary intermediate tensor and incurs additional HBM traffic.

We eliminate this overhead by fusing the broadcast and addition into the epilogue of the candidate-side GEMM. After accumulating each output tile, the epilogue uses the request index associated with each candidate to load the corresponding precomputed request-side contribution. It then performs the addition in registers and writes only the final output to HBM. Consequently, neither the candidate-side GEMM result nor a broadcasted request-side contribution is materialized as an intermediate tensor.

Because candidates belonging to the same request are typically stored consecutively, nearby candidate tiles often access the same request-derived results, improving cache locality and reducing HBM traffic. We implement this fusion as a Triton~\cite{triton} batched-GEMM kernel with a custom post-accumulation epilogue. 

\subsubsection{Overlapping the Mainloop and Epilogue via Warp Specialization}

In a conventional Triton GEMM, software pipelining overlaps operand loads with WGMMA across K blocks within an output tile. Latency outside this pipeline, including the epilogue, can often be hidden by other CTAs resident on the same SM. This mechanism is less effective for our kernel because its large tiles and deep pipeline limit CTA residency. The resulting exposed latency is particularly costly for IKBO, as its epilogue performs an index lookup, loads the corresponding request-side result, and adds it to the accumulated tile rather than simply storing the output. We therefore adopt a persistent, warp-specialized kernel implemented with TLX~\cite{tlx}.

Each persistent CTA repeatedly processes output tiles and is partitioned into three specialized warp groups: one producer and two consumers. The producer continuously issues TMA loads for future tiles, while the consumers operate in a ping-pong manner: one executes the fused epilogue for a completed tile while the other performs WGMMA computation on another tile. This design explicitly overlaps data loading, tensor-core computation, and the fused epilogue without relying on other CTAs on the same SM.

\subsubsection{Combining Request and Candidate Kernels}

The request and candidate computations are still executed as separate kernels, which is inefficient when the request batch is small. In particular, the request-side kernel may expose wave quantization and kernel-launch overhead because it does not contain enough tiles to fully utilize the GPU. We therefore combine the two computations into a single mega-kernel, allowing request- and candidate-side tiles to share a persistent execution schedule. This also creates opportunities to overlap request epilogues with candidate-side input loading.

Candidate-side tiles can be scheduled concurrently with request-side tiles even though their request-side dependencies may not yet be ready. This is possible because the request-side results are consumed only during the candidate-side epilogue. Because the corresponding request and candidate tiles may execute in different CTAs, a candidate epilogue waits on a per-tile completion flag with release-acquire semantics before consuming the request-side result. This avoids a device-wide barrier and allows a candidate tile to proceed as soon as its own dependency is ready.

To improve load balance, we adopt a bidirectional tile scheduling scheme: request-side tiles are assigned in ascending order, while candidate-side tiles are assigned in descending order. This reduces per-SM workload imbalance when the total number of tiles is not evenly divisible by the number of SMs.

\subsubsection{Memory Alignment}
Beyond kernel fusion and scheduling, decomposition also changes the resulting tensor shapes, which can degrade memory-transfer efficiency. In particular, the byte size of a tensor's contiguous dimension may no longer satisfy the alignment requirements for memory transfers and cache transactions. For a row-major tensor, this size determines the byte stride between consecutive rows; a misaligned stride can force narrow loads, prevent use of the TMA path, and make row accesses straddle additional cache lines. We use an automated transformation to identify affected dimensions and zero-pad them, together with the corresponding weight dimensions, according to the element type and backend requirements. This leaves computation mathematically unchanged while enabling wide or TMA transfers and avoiding unnecessary cache-line transactions, thereby reducing L1/TEX pressure and memory traffic.

\subsection{In-Kernel Broadcast Optimization for DCA}
\label{sec:ikbo_dca}

FlashAttention~\cite{dao2022flashattentionfastmemoryefficientexact,flashattention2,flashattention3,flashattention4}
is a widely used optimization for attention workloads. Its efficiency,
however, depends on amortizing each K/V tile load across a sufficiently
large query tile. This amortization is less effective in DCA, where
candidate-side queries are short (e.g., 32 tokens), while request-side
user histories may contain hundreds or thousands of tokens. Each K/V
tile is therefore reused by relatively few query tokens, resulting in
low arithmetic intensity and memory-bound execution.

A straightforward DCA implementation first materializes the projected K/V
tensors at the candidate batch size and then invokes FlashAttention.
Although candidates associated with the same request have identical K/V
values, the materialized copies occupy distinct memory addresses.
FlashAttention consequently treats them as independent batch elements
and cannot reuse the same physical K/V data across candidates.

To better amortize K/V memory access and promote reuse across candidates,
we build on FlashAttention-3~\cite{flashattention3} and specialize its
kernel design for recommendation workloads.

\subsubsection{Amortizing K/V Memory Access across Candidates}

We develop a specialized FlashAttention kernel that directly consumes
the request-side tensors $\mathbf{K}$ and $\mathbf{V}$ at batch
size $B_r$, together with the request--candidate map $\mathbf{m}$.
Before loading a K/V tile for candidate $i$, the kernel resolves $m_i$
and loads from the corresponding request-side K/V tensors, thereby
implementing the logical broadcast inside the kernel. Candidates
associated with the same request consequently access the same physical
K/V data, allowing user-history reads to be amortized across candidates
without materializing candidate-batched copies.

For multi-head attention, we further change the thread-block grid
ordering as follows:
\[
(\texttt{Q\_block}, \texttt{head}, \texttt{candidate})
\;\rightarrow\;
(\texttt{Q\_block}, \texttt{candidate}, \texttt{head}).
\]
Because candidates associated with the same request are stored
consecutively, this ordering schedules blocks that access the same
request-side K/V for a given head closer together, improving temporal
locality and the L2 cache hit rate.

\subsubsection{Persistent Warp-Specialized Execution}
\label{sec:ikbo_dca_persistent}

FlashAttention-3 overlaps loading and tensor-core computation
within a CTA, but the pipeline is drained and reinitialized for each
query tile. This overhead is particularly pronounced for the short
candidate-side queries in DCA. 
We instead use a persistent TLX kernel in which each CTA
processes multiple query tiles. One producer warp group continuously issues
asynchronous TMA loads, while two consumer groups process successive query tiles, each performing WGMMA computation, softmax, and output processing. This overlaps memory loading, tensor-core computation, and output stores while amortizing
buffer and barrier initialization. 

Our WGMMA implementation uses a query-tile size of
$64$ per consumer warp group. We therefore co-process
candidates associated with the same request to fully utilize CTAs. With two consumer warp groups, each CTA processes two candidates when
the query length is 64 and four candidates when the query length is 32.

\section{Evaluation}
\label{sec:eval}

\begin{table*}[t]

\footnotesize\sffamily
\setlength{\tabcolsep}{.5816666em}

\begin{tabular}{lc@{\hskip 0.26166em}c@{\hskip 0.5166em}c@{\hskip 1.8em}c@{\hskip 0.26166em}c@{\hskip 0.5166em}c@{\hskip 1.8em}c@{\hskip 0.26166em}c@{\hskip -0.19166em}c}
\toprule
&
\multicolumn{3}{c}{\textbf{KuaiVideo X1}}
&
\multicolumn{3}{c}{\textbf{KKBox}}
&
\multicolumn{3}{c}{\textbf{KuaiRand X1}}
\\

&
\begin{tabular}[c]{@{}c@{}}
AUC\\
All (Click/Like/Follow)
\end{tabular}
&
{\color[HTML]{656565}
\begin{tabular}[c]{@{}c@{}}
FLOPs@1\\
($C_R/C_C$)
\end{tabular}}
&
{\color[HTML]{656565}
\begin{tabular}[c]{@{}c@{}}
FLOPs@100\\
(Reduced\%)
\end{tabular}}
&
\begin{tabular}[c]{@{}c@{}}
AUC\\
Replay
\end{tabular}
&
{\color[HTML]{656565}
\begin{tabular}[c]{@{}c@{}}
FLOPs@1\\
($C_R/C_C$)
\end{tabular}}
&
{\color[HTML]{656565}
\begin{tabular}[c]{@{}c@{}}
FLOPs@100\\
(Reduced\%)
\end{tabular}}
&
\begin{tabular}[c]{@{}c@{}}
AUC\\
All (Click/Like/Follow)
\end{tabular}
&
{\color[HTML]{656565}
\begin{tabular}[c]{@{}c@{}}
FLOPs@1\\
($C_R/C_C$)
\end{tabular}}
&
\multicolumn{1}{c}{
{\color[HTML]{656565}
\begin{tabular}[c]{@{}c@{}}
FLOPs@100\\
(Reduced\%)
\end{tabular}}
}
\\
\midrule

\addlinespace[2px]

RankMixer
&
0.7933
(0.7199/\textbf{0.7843}/0.8757)
&
{\color[HTML]{656565} 9.52}
&
{\color[HTML]{656565} 9.52}
&
0.8167
&
{\color[HTML]{656565} 5.49}
&
{\color[HTML]{656565} 5.49}
&
0.8635
(\textbf{0.7765}/0.8810/0.9328)
&
{\color[HTML]{656565} 10.59}
&
{\color[HTML]{656565} 10.59}
\\

\hspace{0.516em} UGSEP-Base
&
0.7919
(\textbf{0.7209}/0.7792/0.8756)
&
{\color[HTML]{656565} 5.37/4.74}
&
{\color[HTML]{656565} \underline{4.79 (49.5\%)}}
&
0.8135
&
{\color[HTML]{656565} 3.05/2.77}
&
{\color[HTML]{656565} \underline{2.79 (49.1\%)}}
&
0.8628
(0.7757/0.8789/\textbf{0.9338})
&
{\color[HTML]{656565} 5.68/5.49}
&
{\color[HTML]{656565} \underline{5.51 (47.9\%)}}
\\

\hspace{0.516em} UGSEP-Scaled
&
\textbf{0.7939}
(0.7186/0.7829/\textbf{0.8802})
&
{\color[HTML]{656565} 10.74/9.50}
&
{\color[HTML]{656565} 9.61}
&
\textbf{0.8182}
&
{\color[HTML]{656565} 6.32/5.69}
&
{\color[HTML]{656565} 5.76}
&
\textbf{0.8641}
(0.7763/\textbf{0.8842}/0.9318)
&
{\color[HTML]{656565} 11.19/10.56}
&
{\color[HTML]{656565} 10.67}
\\

\addlinespace[2px]
\hdashline
\addlinespace[4px]

DCNv2 & 0.7909 (\textbf{0.7200}/0.7728/0.8622) & {\color[HTML]{656565} 1.49} & {\color[HTML]{656565} 1.49} & 0.8287 & {\color[HTML]{656565} 7.21} & {\color[HTML]{656565} 7.21} & 0.8692 (0.7755/0.9013/\textbf{0.9316}) & {\color[HTML]{656565} 10.58} & {\color[HTML]{656565} 10.58} \\ \quad ROCS-Base & 0.7850 (0.7156/0.7927/0.8643) & {\color[HTML]{656565} 1.06/0.13} & {\color[HTML]{656565} \underline{0.14 (90.6\%)}} & 0.8250 & {\color[HTML]{656565} 2.66/2.88} & {\color[HTML]{656565} \underline{2.91 (50.7\%)}} & 0.8672 (0.7749/0.8971/0.9289) & {\color[HTML]{656565} 3.49/4.76} & {\color[HTML]{656565} \underline{4.79 (54.7\%)}} \\ \quad ROCS-Scaled & \textbf{0.7935} (0.7138/\textbf{0.7947}/\textbf{0.8719}) & {\color[HTML]{656565} 11.58/1.34} & {\color[HTML]{656565} 1.46} & \textbf{0.8330} & {\color[HTML]{656565} 5.29/5.74} & {\color[HTML]{656565} 5.79} & \textbf{0.8695} (\textbf{0.7770}/\textbf{0.9017}/0.9296) & {\color[HTML]{656565} 7.56/10.31} & {\color[HTML]{656565} 10.38} \\
\addlinespace[2px]
\hdashline
\addlinespace[4px]

FinalMLP
&
0.7848
(0.7158/0.7734/0.8654)
&
{\color[HTML]{656565} 4.39}
&
{\color[HTML]{656565} 4.39}
&
0.8232
&
{\color[HTML]{656565} 10.21}
&
{\color[HTML]{656565} 10.21}
&
0.8661
(0.7750/0.8805/0.9310)
&
{\color[HTML]{656565} 9.96}
&
{\color[HTML]{656565} 9.96}
\\

\quad ROCS-Base
&
0.7843
(0.7163/0.7724/0.8642)
&
{\color[HTML]{656565} 2.24/1.83}
&
{\color[HTML]{656565} \underline{1.87 (57.5\%)}}
&
0.8192
&
{\color[HTML]{656565} 2.75/5.61}
&
{\color[HTML]{656565} \underline{5.64 (44.7\%)}}
&
0.8659
(0.7746/0.8926/0.9305)
&
{\color[HTML]{656565} 2.30/5.23}
&
{\color[HTML]{656565} \underline{5.26 (30.1\%)}}
\\

\quad ROCS-Scaled
&
\textbf{0.7890}
(\textbf{0.7164}/\textbf{0.7819}/\textbf{0.8687})
&
{\color[HTML]{656565} 5.37/4.17}
&
{\color[HTML]{656565} 4.23}
&
\textbf{0.8239}
&
{\color[HTML]{656565} 4.54/11.21}
&
{\color[HTML]{656565} 11.26}
&
\textbf{0.8700}
(\textbf{0.7762}/\textbf{0.9019}/\textbf{0.9319})
&
{\color[HTML]{656565} 5.66/6.47}
&
{\color[HTML]{656565} 6.53}
\\

\addlinespace[2px]
\hdashline
\addlinespace[4px]

Wukong
&
0.7972
(\textbf{0.7247}/0.7918/0.8750)
&
{\color[HTML]{656565} 1.10}
&
{\color[HTML]{656565} 1.10}
&
0.8342
&
{\color[HTML]{656565} 6.93}
&
{\color[HTML]{656565} 6.93}
&
0.8673
(0.7820/0.8838/0.9363)
&
{\color[HTML]{656565} 9.92}
&
{\color[HTML]{656565} 9.92}
\\

\quad ROCS-Base
&
0.7929
(0.7166/0.7803/\textbf{0.8819})
&
{\color[HTML]{656565} 0.83/0.26}
&
{\color[HTML]{656565} \underline{0.27 (75.4\%)}}
&
0.8326
&
{\color[HTML]{656565} 3.07/3.27}
&
{\color[HTML]{656565} \underline{3.30 (53.8\%)}}
&
0.8624
(0.7832/0.8669/0.9370)
&
{\color[HTML]{656565} 3.05/5.44}
&
{\color[HTML]{656565} \underline{5.47 (45.1\%)}}
\\

\quad ROCS-Scaled
&
\textbf{0.7980}
(0.7211/\textbf{0.7928}/0.8801)
&
{\color[HTML]{656565} 3.96/1.05}
&
{\color[HTML]{656565} 1.10}
&
\textbf{0.8357}
&
{\color[HTML]{656565} 7.36/6.78}
&
{\color[HTML]{656565} 6.86}
&
\textbf{0.8715}
(\textbf{0.7847}/\textbf{0.8908}/\textbf{0.9390})
&
{\color[HTML]{656565} 7.24/9.87}
&
{\color[HTML]{656565} 9.92}
\\

\bottomrule
\end{tabular}

\caption{
Results on three public datasets.
The best AUC for each task within each backbone is highlighted in
\textbf{bold}, while the lowest FLOPs@100 is \underline{underlined}.
}

\label{tab:publicdata_results}

\end{table*}

We evaluate ROCS from both model and system perspectives. Our experiments are organized into three parts. 
First, we evaluate ROCS on three public datasets and multiple interaction backbones to study its generality and quality–efficiency frontier. 
Second, we conduct in-depth experiments on multiple production-scale internal datasets using Wukong~\cite{zhang2024wukong} as a representative backbone, evaluating the effectiveness of ROCS end to end and isolating the contribution of its individual components. 
Finally, we report online deployment results across retrieval and ranking systems, demonstrating how ROCS improves model quality, serving efficiency, or both.

\subsection{Public Benchmarks}
\label{sec:eval:public}

We first evaluate whether ROCS generalizes across recommendation datasets and backbones. The public experiments focus on the model-level quality--efficiency frontier exposed by ROCS.

\subsubsection{Evaluation Setup}\leavevmode\\
\para{Datasets}
We select public recommendation benchmarks that provide distinct request- and candidate-side raw features, with neither side represented solely by a single ID. 
This setting better reflects production recommendation workloads and ensures that ROCS has nontrivial request-side computation to expose and reuse.
We use KuaiRand~\cite{gao2022kuairand}, KuaiVideo~\cite{kuaivideo}, and KKBox~\cite{kkbox}, following the preprocessing and data formats provided by RecZoo~\cite{zhu2021open,zhu2022bars}.
Dataset statistics are reported in Table~\ref{tab:public_metadata} in Appendix~\ref{app:public_dataset}.

\para{Metrics}
We evaluate each model in terms of prediction quality and inference complexity.
For prediction quality, we report the area under the ROC curve (AUC) for each feedback type and the macro-average AUC across feedback types.
We report FLOPs to quantify the hardware-independent algorithmic savings enabled by ROCS.
Let $C_R$ denote the request-side FLOPs evaluated once per request and $C_C$ denote the candidate-side FLOPs evaluated once per candidate.
For a request associated with $N$ candidates, the amortized FLOPs per candidate are $C_{\mathrm{amort}}(N)=C_C+C_R/N$.
We report results for $N\in\{1,100\}$, ranging from single-candidate scoring to serving each request against a large candidate set.

\para{Backbones}
We evaluate ROCS on three representative interaction backbones: DCNv2~\cite{wang2021dcn}, FinalMLP~\cite{mao2023finalmlp}, and Wukong~\cite{zhang2024wukong}. 
These models use different feature-interaction modules, allowing us to evaluate whether ROCS generalizes beyond a particular architecture.
We additionally include RankMixer with UGSEP~\cite{lu2026computeonceugseparationefficient} as a request-sharing baseline. Implementation details, hyperparameter choices, and tuning procedures are provided in Appendix~\ref{app:public_implementation}.

\subsubsection{Evaluation Scenarios} For each ROCS backbone, we evaluate the following configurations.

\para{Vanilla}
The original backbone performs early interaction between request- and candidate-side features.
Its intermediate request-side representations therefore become candidate-dependent, requiring the complete model to be evaluated for every candidate.

\para{ROCS-Base}
We configure ROCS to approximately match the inference FLOPs of the corresponding Vanilla model when $N=1$.
This efficiency-oriented configuration does not reinvest the savings available at larger candidate multiplicities, allowing us to isolate the quality effect of directly ROCSifying an early-fusion model.
When exact FLOP matching is not possible because model dimensions are discrete, we select the closest setting within the target budget.

\para{ROCS-Scaled}
We scale ROCS such that its amortized inference FLOPs at $N=100$ approximately match those of the corresponding Vanilla model.
This setting evaluates request-oriented resource reallocation by reinvesting the computation saved through request-side sharing into additional model capacity.

For RankMixer, we report the UGSEP@1 and UGSEP@100 configurations using the same amortized-cost definitions.

\subsubsection{Results} Table~\ref{tab:publicdata_results} reports prediction quality and amortized inference complexity across the evaluated datasets and backbones.

\para{Direct ROCSification favors efficiency.}
Under the base configuration, ROCS substantially reduces amortized FLOPs at $N=100$ across all evaluated ROCS backbones.
As expected, directly ROCSifying a model without scaling may incur a modest quality loss due to the reduced capacity of the candidate-dependent pathway.

\para{RRR improves the quality-efficiency frontier.}
ROCS--Scaled reinvests the saved computation into additional request-side capacity and improves the aggregate AUC of the corresponding Vanilla backbone under a comparable $N=100$ amortized budget.
ROCS can therefore either retain the savings as lower serving cost or reinvest them into model capacity, depending on the objective.

\para{Comparison with specialized request sharing.}
RankMixer with UGSEP serves as a closely related architecture-specific baseline.
Across the evaluated datasets, ROCS enables DCNv2, FinalMLP, and Wukong to achieve competitive quality--efficiency frontiers while supporting substantially different feature-interaction modules.

\subsection{Production-Scale Evaluation}
\label{sec:eval:internal}

The public benchmarks establish the applicability of ROCS across multiple backbones.
We next fix the backbone to Wukong~\cite{zhang2024wukong}, a representative large-scale recommendation backbone, and conduct controlled studies on workloads spanning retrieval and ranking stages and a wide range of model complexities.

\subsubsection{Evaluation Setup}\leavevmode\\
\para{Workloads}
We evaluate three production workloads spanning ads retrieval, short-form video ranking, and ads ranking. Each dataset contains over 100 billion training examples, and the corresponding models span approximately two orders of magnitude in inference complexity.

\para{Baseline}
For each workload, we compare ROCS with the corresponding Vanilla Wukong model under the same data, feature sets, training procedure, and serving environment.

\para{Metrics}
We report relative log-loss (RLL) improvement, amortized inference complexity in MFLOPs per candidate, and relative QPS measured on H100 servers using replayed production traffic.
Lower RLL and higher QPS are better.
RLL improvement of 0.02\% and 0.1\% are considered practically meaningful for the ads and organic workloads, respectively.
FLOPs quantify the model-level efficiency gains enabled by ROCS, whereas QPS captures the end-to-end serving efficiency realized by the ROCS implementation, including IKBO.

\subsubsection{End-to-End Results}

Table~\ref{tab:internal_e2e} reports end-to-end results across the three production workloads.
ROCS matches or improves model quality while increasing replay QPS by 47--196\% across retrieval and ranking models spanning approximately two orders of magnitude in inference complexity.
The largest gain is observed in ads retrieval, which also has the highest candidate multiplicity, consistent with greater amortization of request-side computation.
For short-form video ranking, part of the recovered serving budget is reinvested into model capacity, yielding RLL improvement while retaining a substantial QPS gain.

\begin{table}[h]
\centering
\footnotesize\sffamily
\setlength{\tabcolsep}{0.82em}
\begin{tabular}{llrrrr}
\toprule
\multirow{2}{*}{\textbf{Stage}}
& \multirow{2}{*}{\textbf{Surface}}
& \textbf{Relative}
& \textbf{Complexity}
& \textbf{Candidate}
& \textbf{Relative} \\
& & \textbf{LogLoss}
& \textit{MFLOPs}
& \textbf{Multiplicity}
& \textbf{QPS} \\
\midrule

Retrieval
& Ads
& On par
& $O(100)$
& $O(10{,}000)$
& $+196\%$ \\

\addlinespace[2px]

Ranking
& Organic
& $-0.5\%$
& $O(1{,}000)$
& $O(100)$
& $+47\%$ \\

\addlinespace[2px]

Ranking
& Ads
& On par
& $O(10{,}000)$
& $O(100)$
& $+62\%$ \\

\bottomrule
\end{tabular}
\caption{ROCS performance across deployment stages and surfaces.}
\label{tab:internal_e2e}
\end{table}

\subsubsection{Request-Side Scaling}
\begin{figure}[h]
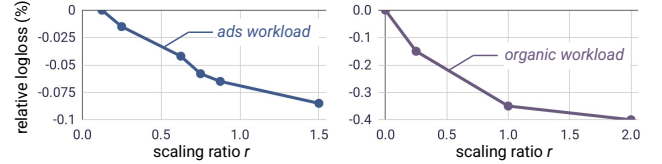

\centering 

\rocsfigureonecol{8}{0.9in}
\vspace{-0.3in}
\caption{RLL improvement increases as request-side computation scales.}
\label{fig:ablation_scale} 
\end{figure}

We fix the candidate-side capacity and progressively scale the request-side computation.
Figure~\ref{fig:ablation_scale} shows that model quality improves consistently as the request-to-candidate FLOPs ratio increases across two models, demonstrating the effectiveness of Request-oriented Resource Reallocation.

\subsubsection{Effectiveness of DCA}

Table~\ref{tab:dca} evaluates three key design choices of DCA.
Replacing DCA with Interformer, which introduces candidate dependence into the sequence-processing stack, achieves comparable quality but reduces QPS by 36\%.
Restricting cross-attention to a single interaction depth degrades RLL by 0.05\%, showing the quality benefit of layer-wise sequence retrieval.
Removing request-side cross-attention causes a 0.05\% RLL regression,  showing that request-side retrieval also contributes to quality.

\begin{table}[h]
\centering
\footnotesize\sffamily
\setlength{\tabcolsep}{0.66em}
\begin{tabular}{lrr}
\toprule
\textbf{Model}
& \textbf{Relative LogLoss}
& \textbf{Relative QPS} \\
\midrule

DCA
& (baseline)
& (baseline) \\

\addlinespace[2px]
\hdashline
\addlinespace[4px]

Interformer
& On par
& $-36\%$ \\

Single-Depth Cross-Attention
& $+0.05\%$
& $+5\%$ \\

DCA w/o Request-side Cross Attention
& $+0.05\%$
& On par \\

\bottomrule
\end{tabular}
\caption{
Ablation of the key design choices in DCA.
}
\label{tab:dca}
\end{table}

\subsubsection{Effectiveness of IKBO}

Table~\ref{tab:ikbo_lcb} reports the incremental optimization of the IKBO-LCB kernel.
Fusing request-side broadcast into the candidate GEMM and applying the subsequent scheduling and alignment optimizations progressively reduce kernel latency.

\begin{table}[h]
\centering
\footnotesize\sffamily%
\setlength{\tabcolsep}{1.36em}
\begin{tabular}{lrr}
\toprule
\textbf{Optimization} 
    & \textbf{Latency (ms)}
    & \textbf{Changes} \\
\midrule

Original
    & 1.944
    & (baseline) \\

\addlinespace[2px]
\hdashline
\addlinespace[4px]

Decomposition
    & 1.389
    & $-28.5\%$ \\

Memory alignment
    & 0.798
    & $-58.9\%$ \\

Broadcast fusion
    & 0.580
    & $-70.2\%$ \\

Pipelining GEMM with broadcast
    & 
    & \\\quad + fusing request/candidate compute
& 0.482
    & $-75.2\%$ \\

\bottomrule
\end{tabular}

\caption{
Latency breakdown of incremental IKBO-LCB kernel.
}
\label{tab:ikbo_lcb}
\end{table}

Table~\ref{tab:ikbo_dca} compares IKBO-DCA with optimized FlashAttention baselines.
IKBO reduces attention latency from 0.55\,ms to 0.23\,ms by eliminating materialized K/V broadcast and improving reuse across candidates.
Measured on the attention operator, IKBO reduces memory bandwidth while increasing throughput, shifting the kernel from memory-bound toward compute-bound.

\begin{table}[h]
\centering
\footnotesize\sffamily
\setlength{\tabcolsep}{0.8em}
\begin{tabular}{lrrr}
\toprule
\textbf{Method}
& \textbf{Throughput}
& \textbf{IO}
& \textbf{Total / Attention Latency} \\
& \textit{TFLOPs/s}
& \textit{GB/s}
& \textit{ms} \\
\midrule

TLX FA3
& 245
& 2,152
& 1.473 / 0.561 \\

CuTeDSL FA4 Hopper
& 250
& 2,193
& 1.462 / 0.550 \\

\addlinespace[2px]
\hdashline
\addlinespace[4px]

\textbf{IKBO DCA}
& \textbf{594}
& \textbf{681}
& \textbf{0.230 / 0.230} \\

\bottomrule
\end{tabular}

\caption{
Performance comparison of IKBO-DCA and FlashAttention baselines.
Total latency includes explicit K/V broadcast and attention.
}
\label{tab:ikbo_dca}
\end{table}

\subsection{Online Deployment}

\rocs has been deployed across tens of production recommendation models at Meta, spanning ads and organic surfaces at both retrieval and ranking stages. Representative results include: (1) 0.04\% topline metric gain with 32\% capacity savings on a short-form video ranking model, a practically significant improvement at this scale; (2) on-par topline with 38\% capacity savings on an ads ranking model; and (3) 0.2\% topline metric gain with 29\% capacity savings on an ads retrieval model. Across deployments, \rocs consistently delivers quality gains, cost reductions, or both, and the freed capacity can be reinvested into further model scaling.

\section{Discussion and Conclusion}
ROCS is most effective when each request is evaluated against multiple candidates and a substantial fraction of model computation remains candidate-independent. Its benefits diminish when candidate multiplicity is low, including in low-candidate training settings.
While IKBO is a general principle, our implementation relies on primitives available on modern NVIDIA GPUs.

We presented ROCS, a modeling and inference paradigm that preserves a
reusable request-side pathway while retaining intermediate
request--candidate interaction. GLM, DCA, and RRR expose and reinvest
request-shared computation, while IKBO realizes this sharing efficiently
without materialized broadcast. Across public benchmarks,
production-scale workloads, and online deployments, ROCS 
improves the quality--efficiency frontier. These results demonstrate
that request-level redundancy is a structural model-system codesign
opportunity, rather than merely an implementation artifact.

\bibliographystyle{ACM-Reference-Format}
\bibliography{rocs}

@inproceedings{cdeep,
  title={Deep neural networks for {YouTube} recommendations},
  author={Covington, Paul and Adams, Jay and Sargin, Emre},
  booktitle={Proceedings of the 10th ACM Conference on Recommender Systems},
  pages={191--198},
  year={2016}
}

@inproceedings{lattice,
  title={Meta lattice: Model space redesign for cost-effective industry-scale ads recommendations},
  author={Luo, Liang and Chen, Yuxin and Zhang, Zhengyu and Hang, Mengyue and Gu, Andrew and Zhang, Buyun and Liu, Boyang and Chen, Chen and Yang, Fan and Gu, Feifan and others},
  booktitle={Proceedings of the 32nd ACM SIGKDD Conference on Knowledge Discovery and Data Mining V. 1},
  pages={2335--2346},
  year={2026}
}

@article{solaris,
  title={SOLARIS: Speculative Offloading of Latent-bAsed Representation for Inference Scaling},
  author={Liu, Zikun and Luo, Liang and Li, Qianru and Zhang, Zhengyu and Ling, Wei and Shen, Jingyi and Chen, Zeliang and Huang, Yaning and Huang, Jingxian and Aboelela, Abdallah and others},
  journal={arXiv preprint arXiv:2604.12110},
  year={2026}
}

@article{exlargefm,
  title={External Large Foundation Model: How to Efficiently Serve Trillions of Parameters for Online Ads Recommendation},
  author={Recommendation, Ads},
  journal={arXiv preprint arXiv:2502.17494},
  year={2025}
}

@inproceedings{zhu2022bars,
  title={Bars: Towards open benchmarking for recommender systems},
  author={Zhu, Jieming and Dai, Quanyu and Su, Liangcai and Ma, Rong and Liu, Jinyang and Cai, Guohao and Xiao, Xi and Zhang, Rui},
  booktitle={Proceedings of the 45th International ACM SIGIR Conference on Research and Development in Information Retrieval},
  pages={2912--2923},
  year={2022}
}

@inproceedings{gao2022kuairand,
  title={Kuairand: An unbiased sequential recommendation dataset with randomly exposed videos},
  author={Gao, Chongming and Li, Shijun and Zhang, Yuan and Chen, Jiawei and Li, Biao and Lei, Wenqiang and Jiang, Peng and He, Xiangnan},
  booktitle={Proceedings of the 31st ACM international conference on information \& knowledge management},
  pages={3953--3957},
  year={2022}
}

@inproceedings{zhu2021open,
  title={Open benchmarking for click-through rate prediction},
  author={Zhu, Jieming and Liu, Jinyang and Yang, Shuai and Zhang, Qi and He, Xiuqiang},
  booktitle={Proceedings of the 30th ACM international conference on information \& knowledge management},
  pages={2759--2769},
  year={2021}
}

@inproceedings{mao2023finalmlp,
  title={FinalMLP: an enhanced two-stream MLP model for CTR prediction},
  author={Mao, Kelong and Zhu, Jieming and Su, Liangcai and Cai, Guohao and Li, Yuru and Dong, Zhenhua},
  booktitle={Proceedings of the AAAI conference on artificial intelligence},
  volume={37},
  number={4},
  pages={4552--4560},
  year={2023}
}

@inproceedings{zhu2025rankmixer,
  title={Rankmixer: Scaling up ranking models in industrial recommenders},
  author={Zhu, Jie and Fan, Zhifang and Zhu, Xiaoxie and Jiang, Yuchen and Wang, Hangyu and Han, Xintian and Ding, Haoran and Wang, Xinmin and Zhao, Wenlin and Gong, Zhen and others},
  booktitle={Proceedings of the 34th ACM International Conference on Information and Knowledge Management},
  pages={6309--6316},
  year={2025}
}

@inproceedings{zhang2024wukong,
  title={Wukong: towards a scaling law for large-scale recommendation},
  author={Zhang, Buyun and Luo, Liang and Chen, Yuxin and Nie, Jade and Liu, Xi and Li, Shen and Zhao, Yanli and Hao, Yuchen and Yao, Yantao and Wen, Ellie Dingqiao and others},
  booktitle={Proceedings of the 41st International Conference on Machine Learning},
  pages={59421--59434},
  year={2024}
}

@misc{lu2026computeonceugseparationefficient,
      title={Compute Only Once: UG-Separation for Efficient Large Recommendation Models}, 
      author={Hui Lu and Zheng Chai and Shipeng Bai and Hao Zhang and Zhifang Fan and Kunmin Bai and Ke Sun and Yingwen Wu and Bingzheng Wei and Xiang Sun and Ziyan Gong and Tianyi Liu and Hua Chen and Deping Xie and Zhongkai Chen and Zhiliang Guo and Qiwei Chen and Yuchao Zheng},
      year={2026},
      eprint={2602.10455},
      archivePrefix={arXiv},
      primaryClass={cs.IR},
      url={https://arxiv.org/abs/2602.10455}, 
}

@article{dlrm,
  title={Deep learning recommendation model for personalization and recommendation systems},
  author={Naumov, Maxim and Mudigere, Dheevatsa and Shi, Hao-Jun Michael and Huang, Jianyu and Sundaraman, Narayanan and Park, Jongsoo and Wang, Xiaodong and Gupta, Udit and Wu, Carole-Jean and Azzolini, Alisson G and others},
  journal={arXiv preprint arXiv:1906.00091},
  year={2019}
}

@inproceedings{wang2021dcn,
  title={Dcn v2: Improved deep \& cross network and practical lessons for web-scale learning to rank systems},
  author={Wang, Ruoxi and Shivanna, Rakesh and Cheng, Derek and Jain, Sagar and Lin, Dong and Hong, Lichan and Chi, Ed},
  booktitle={Proceedings of the web conference 2021},
  pages={1785--1797},
  year={2021}
}

@article{deck,
  title={DECK: Experiences on Delta Checkpointing for Industrial Recommendation Systems},
  author={Gao, Xin and Acharya, Sibasish and Han, Sihui and Ren, Yongxiong and Zhao, Yanli and Luo, Liang and Wang, Chucheng and Fernando, Pradeep and Mishra, Saurabh and Yan, Siqi and others},
  journal={Proceedings of the VLDB Endowment},
  volume={18},
  number={12},
  pages={4978--4990},
  year={2025},
  publisher={VLDB Endowment}
}

@article{zhang2022dhen,
  title={DHEN: A deep and hierarchical ensemble network for large-scale click-through rate prediction},
  author={Zhang, Buyun and Luo, Liang and Liu, Xi and Li, Jay and Chen, Zeliang and Zhang, Weilin and Wei, Xiaohan and Hao, Yuchen and Tsang, Michael and Wang, Wenjun and others},
  journal={arXiv preprint arXiv:2203.11014},
  year={2022}
}

@article{dmt,
  title={Disaggregated multi-tower: Topology-aware modeling technique for efficient large scale recommendation},
  author={Luo, Liang and Zhang, Buyun and Tsang, Michael and Ma, Yinbin and Chen, Yuxin and Chu, Ching-Hsiang and Zhao, Yanli and Li, Shen and Hao, Yuchen and Lakshminarayanan, Guna and others},
  journal={Proceedings of Machine Learning and Systems},
  volume={6},
  pages={266--278},
  year={2024}
}

@inproceedings{zeng2025interformer,
  title={InterFormer: Effective Heterogeneous Interaction Learning for Click-Through Rate Prediction},
  author={Zeng, Zhichen and Liu, Xiaolong and Hang, Mengyue and Liu, Xiaoyi and Zhou, Qinghai and Yang, Chaofei and Liu, Yiqun and Ruan, Yichen and Chen, Laming and Chen, Yuxin and others},
  booktitle={Proceedings of the 34th ACM International Conference on Information and Knowledge Management},
  pages={6225--6233},
  year={2025}
}

@article{hou2026kunlun,
  title={Kunlun: Establishing scaling laws for massive-scale recommendation systems through unified architecture design},
  author={Hou, Bojian and Liu, Xiaolong and Liu, Xiaoyi and Xu, Jiaqi and Badr, Yasmine and Hang, Mengyue and Chanpuriya, Sudhanshu and Zhou, Junqing and Yang, Yuhang and Xu, Han and others},
  journal={arXiv preprint arXiv:2602.10016},
  year={2026}
}

@article{vaswani2017attention,
  title={Attention is all you need},
  author={Vaswani, Ashish and Shazeer, Noam and Parmar, Niki and Uszkoreit, Jakob and Jones, Llion and Gomez, Aidan N and Kaiser, {\L}ukasz and Polosukhin, Illia},
  journal={Advances in neural information processing systems},
  volume={30},
  year={2017}
}

@misc{dao2022flashattentionfastmemoryefficientexact,
      title={FlashAttention: Fast and Memory-Efficient Exact Attention with IO-Awareness}, 
      author={Tri Dao and Daniel Y. Fu and Stefano Ermon and Atri Rudra and Christopher Ré},
      year={2022},
      eprint={2205.14135},
      archivePrefix={arXiv},
      primaryClass={cs.LG},
      url={https://arxiv.org/abs/2205.14135}, 
}

@article{kaplan2020scaling,
  title={Scaling laws for neural language models},
  author={Kaplan, Jared and McCandlish, Sam and Henighan, Tom and Brown, Tom B and Chess, Benjamin and Child, Rewon and Gray, Scott and Radford, Alec and Wu, Jeffrey and Amodei, Dario},
  journal={arXiv preprint arXiv:2001.08361},
  year={2020}
}

@article{ardalani2022understanding,
  title={Understanding scaling laws for recommendation models},
  author={Ardalani, Newsha and Wu, Carole-Jean and Chen, Zeliang and Bhushanam, Bhargav and Aziz, Adnan},
  journal={arXiv preprint arXiv:2208.08489},
  year={2022}
}

@article{guo2023embedding,
  title={On the Embedding Collapse when Scaling up Recommendation Models},
  author={Guo, Xingzhuo and Pan, Junwei and Wang, Ximei and Chen, Baixu and Jiang, Jie and Long, Mingsheng},
  journal={arXiv preprint arXiv:2310.04400},
  year={2023}
}

@misc{luo2026lokalowprecisionkernelapplications,
      title={LoKA: Low-precision Kernel Applications for Recommendation Models At Scale}, 
      author={Liang Luo and Yinbin Ma and Quanyu Zhu and Vasiliy Kuznetsov and Yuxin Chen and Neng Shi and Jian Jiao and Jiecao Yu and Buyun Zhang and Tongyi Tang and Xiaohan Wei and Yanli Zhao and Zeliang Chen and Yuchen Hao and Venkatesh Ranganathan and Sandeep Parab and Yantao Yao and Maxim Naumov and Chunzhi Yang and Shen Li and Ellie Wen and Wenlin Chen and Santanu Kolay and Chunqiang Tang},
      year={2026},
      eprint={2605.10886},
      archivePrefix={arXiv},
      primaryClass={cs.LG},
      url={https://arxiv.org/abs/2605.10886}, 
}

@inproceedings{zhou2018din,
  title={Deep interest network for click-through rate prediction},
  author={Zhou, Guorui and Zhu, Xiaoqiang and Song, Chenru and Fan, Ying and Zhu, Han and Ma, Xiao and Yan, Yanghui and Jin, Junqi and Li, Han and Gai, Kun},
  booktitle={Proceedings of the 24th ACM SIGKDD international conference on knowledge discovery \& data mining},
  pages={1059--1068},
  year={2018}
}

@inproceedings{pi2020sim,
  title={Search-based user interest modeling with lifelong sequential behavior data for click-through rate prediction},
  author={Pi, Qi and Zhou, Guorui and Zhang, Yujing and Wang, Zhe and Ren, Lejian and Fan, Ying and Zhu, Xiaoqiang and Gai, Kun},
  booktitle={Proceedings of the 29th ACM International Conference on Information \& Knowledge Management},
  pages={2685--2692},
  year={2020}
}

@inproceedings{chang2023twin,
  title={TWIN: TWo-stage interest network for lifelong user behavior modeling in CTR prediction at kuaishou},
  author={Chang, Jianxin and Zhang, Chenbin and Fu, Zhiyi and Zang, Xiaoxue and Guan, Lin and Lu, Jing and Hui, Yiqun and Leng, Dewei and Niu, Yanan and Song, Yang and others},
  booktitle={Proceedings of the 29th ACM SIGKDD Conference on Knowledge Discovery and Data Mining},
  pages={3785--3794},
  year={2023}
}

@inproceedings{si2024twinv2,
  title={Twin v2: Scaling ultra-long user behavior sequence modeling for enhanced ctr prediction at kuaishou},
  author={Si, Zihua and Guan, Lin and Sun, ZhongXiang and Zang, Xiaoxue and Lu, Jing and Hui, Yiqun and Cao, Xingchao and Yang, Zeyu and Zheng, Yichen and Leng, Dewei and others},
  booktitle={Proceedings of the 33rd ACM International Conference on Information and Knowledge Management},
  pages={4890--4897},
  year={2024}
}

@inproceedings{chai2025longer,
  title={Longer: Scaling up long sequence modeling in industrial recommenders},
  author={Chai, Zheng and Ren, Qin and Xiao, Xijun and Yang, Huizhi and Han, Bo and Zhang, Sijun and Chen, Di and Lu, Hui and Zhao, Wenlin and Yu, Lele and others},
  booktitle={Proceedings of the Nineteenth ACM Conference on Recommender Systems},
  pages={247--256},
  year={2025}
}

@misc{kuaivideo,
  title        = {{KuaiVideo}},
  howpublished = {\url{https://www.kuaishou.com/activity/uimc}},
  year         = {2022},
}

@misc{kkbox,
  title        = {{KKBox Music Recommendation Challenge}},
  howpublished = {\url{https://www.kaggle.com/c/kkbox-music-recommendation-challenge}},
  year         = {2018},
}

@inproceedings{triton,
author = {Tillet, Philippe and Kung, H. T. and Cox, David},
title = {Triton: an intermediate language and compiler for tiled neural network computations},
year = {2019},
isbn = {9781450367196},
publisher = {Association for Computing Machinery},
address = {New York, NY, USA},
url = {https://doi.org/10.1145/3315508.3329973},
doi = {10.1145/3315508.3329973},
booktitle = {Proceedings of the 3rd ACM SIGPLAN International Workshop on Machine Learning and Programming Languages},
pages = {10–19},
numpages = {10},
location = {Phoenix, AZ, USA},
series = {MAPL 2019}
}

@article{tlx,
      title={TLX: Hardware-Native, Evolvable MIMW GPU Compiler for Large-scale Production Environments}, 
      author={Yue Guan and Hongtao Yu and Peng Chen and Daohang Shi and Karthik Manivannan and Nicholas J Riasanovsky and Manman Ren and Lei Wang and Shane Nay and Partha Kanuparthy and Zaifeng Pan and Zhengding Hu and Yufei Ding},
      year={2026},
      eprint={2605.10905},
      archivePrefix={arXiv},
      primaryClass={cs.AR},
      url={https://arxiv.org/abs/2605.10905}, 
}

@inproceedings{flashattention2,
 author = {Dao, Tri},
 booktitle = {International Conference on Learning Representations},
 editor = {B. Kim and Y. Yue and S. Chaudhuri and K. Fragkiadaki and M. Khan and Y. Sun},
 pages = {35549--35562},
 title = {FlashAttention-2: Faster Attention with Better Parallelism and Work Partitioning},
 url = {https://proceedings.iclr.cc/paper_files/paper/2024/file/98ed250b203d1ac6b24bbcf263e3d4a7-Paper-Conference.pdf},
 volume = {2024},
 year = {2024}
}

@inproceedings{flashattention3,
author = {Shah, Jay and Bikshandi, Ganesh and Zhang, Ying and Thakkar, Vijay and Ramani, Pradeep and Dao, Tri},
title = {FlashAttention-3: fast and accurate attention with asynchrony and low-precision},
year = {2024},
isbn = {9798331314385},
publisher = {Curran Associates Inc.},
address = {Red Hook, NY, USA},
booktitle = {Proceedings of the 38th International Conference on Neural Information Processing Systems},
articleno = {2193},
numpages = {28},
location = {Vancouver, BC, Canada},
series = {NIPS '24}
}

@misc{flashattention4,
      title={FlashAttention-4: Algorithm and Kernel Pipelining Co-Design for Asymmetric Hardware Scaling}, 
      author={Ted Zadouri and Markus Hoehnerbach and Jay Shah and Timmy Liu and Vijay Thakkar and Tri Dao},
      year={2026},
      eprint={2603.05451},
      archivePrefix={arXiv},
      primaryClass={cs.CL},
      url={https://arxiv.org/abs/2603.05451}, 
}

@misc{guo2025requestonlyoptimizationrecommendationsystems,
      title={Request-Only Optimization for Recommendation Systems}, 
      author={Liang Guo and Wei Li and Lucy Liao and Huihui Cheng and Rui Zhang and Yu Shi and Yueming Wang and Yanzun Huang and Keke Zhai and Pengchao Wang and Timothy Shi and Xuan Cao and Shengzhi Wang and Renqin Cai and Zhaojie Gong and Omkar Vichare and Rui Jian and Leon Gao and Shiyan Deng and Xingyu Liu and Xiong Zhang and Fu Li and Wenlei Xie and Bin Wen and Rui Li and Lu Fang and Xing Liu and Jiaqi Zhai},
      year={2025},
      eprint={2508.05640},
      archivePrefix={arXiv},
      primaryClass={cs.IR},
      url={https://arxiv.org/abs/2508.05640}, 
}

@article{yan2022apg,
  title={Apg: Adaptive parameter generation network for click-through rate prediction},
  author={Yan, Bencheng and Wang, Pengjie and Zhang, Kai and Li, Feng and Deng, Hongbo and Xu, Jian and Zheng, Bo},
  journal={Advances in Neural Information Processing Systems},
  volume={35},
  pages={24740--24752},
  year={2022}
}

@misc{zhang2026onetransunifiedfeatureinteraction,
      title={OneTrans: Unified Feature Interaction and Sequence Modeling with One Transformer in Industrial Recommender}, 
      author={Zhaoqi Zhang and Haolei Pei and Jun Guo and Tianyu Wang and Yufei Feng and Hui Sun and Shaowei Liu and Aixin Sun},
      year={2026},
      eprint={2510.26104},
      archivePrefix={arXiv},
      primaryClass={cs.IR},
      url={https://arxiv.org/abs/2510.26104}, 
}

@misc{huang2026hyformerrevisitingrolessequence,
      title={HyFormer: Revisiting the Roles of Sequence Modeling and Feature Interaction in CTR Prediction}, 
      author={Yunwen Huang and Shiyong Hong and Xijun Xiao and Jinqiu Jin and Xuanyuan Luo and Zhe Wang and Zheng Chai and Shikang Wu and Yuchao Zheng and Jingjian Lin},
      year={2026},
      eprint={2601.12681},
      archivePrefix={arXiv},
      primaryClass={cs.IR},
      url={https://arxiv.org/abs/2601.12681}, 
}

@misc{ding2026ultrahstu,
      title={Bending the Scaling Law Curve in Large-Scale Recommendation Systems}, 
      author={Qin Ding and Kevin Course and Linjian Ma and Jianhui Sun and Ruochen Liu and Zhao Zhu and Chunxing Yin and Wei Li and Dai Li and Yu Shi and Xuan Cao and Ze Yang and Han Li and Xing Liu and Bi Xue and Hongwei Li and Rui Jian and Daisy Shi He and Jing Qian and Matt Ma and Qunshu Zhang and Rui Li},
      year={2026},
      eprint={2602.16986},
      archivePrefix={arXiv},
      primaryClass={cs.IR},
      url={https://arxiv.org/abs/2602.16986}, 
}

@misc{jiang2026tokenmixerlargescalinglargeranking,
      title={TokenMixer-Large: Scaling Up Large Ranking Models in Industrial Recommenders}, 
      author={Yuchen Jiang and Jie Zhu and Xintian Han and Hui Lu and Kunmin Bai and Mingyu Yang and Shikang Wu and Ruihao Zhang and Wenlin Zhao and Shipeng Bai and Sijin Zhou and Huizhi Yang and Tianyi Liu and Wenda Liu and Ziyan Gong and Haoran Ding and Zheng Chai and Deping Xie and Zhe Chen and Yuchao Zheng and Peng Xu},
      year={2026},
      eprint={2602.06563},
      archivePrefix={arXiv},
      primaryClass={cs.IR},
      url={https://arxiv.org/abs/2602.06563}, 
}

@misc{huang2026mixformercoscalingdensesequence,
      title={MixFormer: Co-Scaling Up Dense and Sequence in Industrial Recommenders}, 
      author={Xu Huang and Hao Zhang and Zhifang Fan and Yunwen Huang and Zhuoxing Wei and Zheng Chai and Jinan Ni and Yuchao Zheng and Qiwei Chen},
      year={2026},
      eprint={2602.14110},
      archivePrefix={arXiv},
      primaryClass={cs.IR},
      url={https://arxiv.org/abs/2602.14110}, 
}

@misc{zhai2024actionsspeaklouderwords,
      title={Actions Speak Louder than Words: Trillion-Parameter Sequential Transducers for Generative Recommendations}, 
      author={Jiaqi Zhai and Lucy Liao and Xing Liu and Yueming Wang and Rui Li and Xuan Cao and Leon Gao and Zhaojie Gong and Fangda Gu and Michael He and Yinghai Lu and Yu Shi},
      year={2024},
      eprint={2402.17152},
      archivePrefix={arXiv},
      primaryClass={cs.LG},
      url={https://arxiv.org/abs/2402.17152}, 
}

@article{kingma2014adam,
  title={Adam: A method for stochastic optimization},
  author={Kingma, Diederik P and Ba, Jimmy},
  journal={arXiv preprint arXiv:1412.6980},
  year={2014}
}

@article{paszke2019pytorch,
  title={Pytorch: An imperative style, high-performance deep learning library},
  author={Paszke, Adam and Gross, Sam and Massa, Francisco and Lerer, Adam and Bradbury, James and Chanan, Gregory and Killeen, Trevor and Lin, Zeming and Gimelshein, Natalia and Antiga, Luca and others},
  journal={Advances in neural information processing systems},
  volume={32},
  year={2019}
}

\appendix
\section{Experiment Details}

\subsection{Public Datasets}

\subsubsection{Dataset Details}
\label{app:public_dataset}

The meta-data of selected public datasets are shown in Table.~\ref{tab:public_metadata}.
\begin{table}[h]
\centering
\footnotesize\sffamily
\setlength{\tabcolsep}{0.6em}
\resizebox{\linewidth}{!}{%
\begin{tabular}{lrrrrrl}
\toprule
\textbf{Dataset}
& \textbf{\#Users}
& \textbf{\#Items}
& \textbf{\#Interactions}
& \textbf{\#User Feat.}
& \textbf{\#Item Feat.}
& \textbf{Feedback} \\
\midrule

KuaiRand
& 1,000
& 4,369,953
& 11,713,045
& 30
& 62
& IsClick, IsLike, IsFollow \\

KuaiVideo
& 10,000
& 3,239,534
& 13,661,383
& 5
& 2
& IsClick, IsLike, IsFollow \\

KKBox
& --
& --
& 7,377,418
& 7
& 12
& IsReplay \\

\bottomrule
\end{tabular}%
}

\caption{Metadata of the public datasets.}
\label{tab:public_metadata}
\end{table}

\subsubsection{Implementation Details}
\label{app:public_implementation}
For the experiments on public data, all models are implemented in PyTorch~\cite{paszke2019pytorch} and trained on a single NVIDIA~H100 GPU~(96\,GB).

The hyper-parameters of each vanilla backbone are determined via grid search. The computational cost budget for vanilla backbones is restricted to a forward cost on the order of $10$~MFLOPs per instance. Concretely, for every (model, dataset) pair we draw $20$ configurations from the search grid summarized in Table.~\ref{tab:hp-search-space}, retain those whose cost satisfies the budget, and select the feasible
  configuration attaining the highest validation AUC. All models are trained under an
  identical regime---Adam~\cite{kingma2014adam} with a learning rate of $10^{-3}$, a batch size of $65{,}536$, and up to $100$ epochs with early stopping (patience $5$). For each ROCS-Base configuration, they follow the hyper-parameter choice of their corresponding vanilla backbones. As to the ROCS-Scaled configurations, the capacity is progressively enlarged (e.g., DNN width for DCNv2/FinalMLP) so that the configuration whose amortized per-item cost at $N{=}100$
  does not exceed the vanilla budge.

\newpage
\begin{table}[h]
\centering
\footnotesize\sffamily
\setlength{\tabcolsep}{0.8em}
\resizebox{\linewidth}{!}{%
\begin{tabular}{lll}
\toprule
\textbf{Model}
& \textbf{Hyperparameter}
& \textbf{Search Space} \\
\midrule

\multirow{4}{*}{DCNv2}
& Embedding dimension
& $\{16,\,32,\,64\}$ \\
& \# cross layers
& $\{1,\,2,\,3,\,4\}$ \\
& Deep-network units
& $\{[256,128],\,[400,400],\,[512,256],\,[1024,512,256]\}$ \\
& Network dropout
& $\{0,\,0.1,\,0.3,\,0.5\}$ \\

\addlinespace[2px]
\hdashline
\addlinespace[4px]

\multirow{6}{*}{Wukong}
& Embedding dimension $d$
& $\{64\}$ \\
& \# Wukong layers $l$
& $\{2,\,4,\,8\}$ \\
& Compression dimension $k$
& $\{8,\,16,\,32,\,64\}$ \\
& FMB MLP hidden dimension
& $\{64,\,128,\,256,\,512\}$ \\
& Projection dimension
& $\{64,\,128,\,256\}$ \\
& Prediction-head MLP
& $\{[64],\,[256],\,[512]\}$ \\

\addlinespace[2px]
\hdashline
\addlinespace[4px]

\multirow{6}{*}{FinalMLP}
& Embedding dimension
& $\{16,\,32,\,64\}$ \\
& MLP$_1$ units
& $\{[400,400],\,[400,400,400],\,[400,400,400,400]\}$ \\
& MLP$_2$ units
& $\{[400],\,[1000],\,[2000]\}$ \\
& FS-gate hidden units
& $\{[800]\}$ \\
& \# fusion heads
& $\{200\}$ \\
& MLP dropout
& $\{0.1,\,0.2,\,0.3\}$ \\

\addlinespace[2px]
\hdashline
\addlinespace[4px]

\multirow{6}{*}{RankMixer}
& Embedding dimension
& $\{64\}$ \\
& \# tokens $T$
& $\{4,\,8,\,16\}$ \\
& Token dimension $D$
& $\{64,\,128\}$ \\
& \# layers
& $\{2,\,3,\,4\}$ \\
& Expansion ratio
& $\{2,\,4\}$ \\
& Dropout
& $\{0.1,\,0.2,\,0.3\}$ \\

\bottomrule
\end{tabular}%
}

\caption{Hyperparameter search spaces of the vanilla backbones on public datasets.}
\label{tab:hp-search-space}
\end{table}

\end{document}